# Symmetry Based Cluster Approach for Automatic Recognition of the Epileptic Focus in Brain Using PET Scan Image – An Analysis

- A. Meena[1], K. Raja[2]

[1]*Research Scholar, Dept. of CSE, Sathyabama University, Chennai*
`kabimeena2@hotmail.com`
[2]*Dept. of CSE, Narasu's Sarathy Institute of Technology, Salem*
`raja_koth@yahoo.co.in`

*Abstract*

**Recognition of epileptic focal point is the important diagnosis when screening the epilepsy patients for latent surgical cures. The accurate localization is challenging one because of the low spatial resolution images with more noisy data. Positron Emission Tomography (PET) has now replaced the issues and caring a high resolution. This paper focuses the research of automated localization of epileptic seizures in brain functional images using symmetry based cluster approach. This approach presents a fully automated symmetry based brain abnormality detection method for PET sequences. PET images are spatially normalized to Digital Imaging and Communications in Medicine (DICOM) standard and then it has been trained using symmetry based cluster approach using Medical Image Processing, Analysis & Visualization (MIPAV) tool. The performance evolution is considered by the metric like accuracy of diagnosis. The obtained result is surely assists the surgeon for the automated identification of seizures focus.**

**Key words: Epilepsy, PET scan image, Symmetry based cluster, DICOM standard**

## 1. Introduction

Medical imaging modalities like Computed Tomography (CT), Magnetic Resonance Imaging (MRI) and PET are an integral part of epilepsy diagnosis and providing subsequent treatment planning for effective assessment. The diagnosis of epilepsy and the localization of epileptic focus are made using the criteria weighted by clinical manifestations, electrical patterns identified on the Electroencephalogram (EEG), structural abnormalities on magnetic resonance image (MRI) and functional changes on Positron Emission Tomography (PET) studies [1], [2]. PET is able to detect fine functional changes at the early stages of a disease process, which gives unique feature over other anatomical imaging techniques in the evaluation of neurological disorders, tumors and injury detection, etc in brain [3]. The role of PET has evolved rapidly in the detection of brain abnormalities in human body. A variety of PET image segmentation approaches in brain have proposed in [4], [5].

Epilepsy is the common disease of the Central Nervous System (CNS). Epilepsy people continually face social disgrace and exclusion. Data from ILAE/IBE/WHO reinforce the need for urgent, generous and systematic action to develop resources of epilepsy care. In many cases it is not completely cures but usually controlled with medication, although surgery may be considered in difficult cases [6]. Reduction of interictal cerebral glucose metabolism is one of the characteristic functional abnormalities in epilepsy patient and can be identified using glucose metabolic imaging with F-fluorodeoxyglucose (FDG) and PET [7]. Many literature studies have demonstrated that in approximately 70% of patients with severe partial seizures regional interictal cerebral glucose metabolism is reduced [8].

## 2. Epileptic seizures

Epilepsy is a kind of chronic illness of the brain which can be represented as seizure. The causes of epilepsy can be determined an imbalance of neurotransmitters in the brain that help the nerve cells in the brain transmit electrical impulses. It is considered to be mutations in several genes that have been linked to some types of epilepsy and children born with an abnormal structure of brain. Basal ganglia is a part of brain which formed DOPA a chemical substance. It is necessary for the functional activities of brain. If the level of DOPA is reduced may cause the neurological alterations called Parkinson's disease. During an epileptic seizure, brain nerve cells fire electric impulse much faster than normal. It mainly leads to loss of consciousness and seizures. In early stages, the attack starts identifiable area of the brain, and then broadens to other areas.

According to Theodore pilot study, in United States of America approximately two million adults and 500,000 children have been diagnosed with epilepsy. From that, 600,000 people have seizures that are not controlled by antiepileptic drugs. When drugs fail to control epilepsy, brain surgery is often the only remaining remedial option. But, localizing the epileptic focus is more complicated one. Surgery may be safe and more effective to remove the affected area where attack begins and it is needful to recover from further motor damaging and intellectual function. The fig.1 has shown the various stages of epilepsy occurred by some pathological conditions.

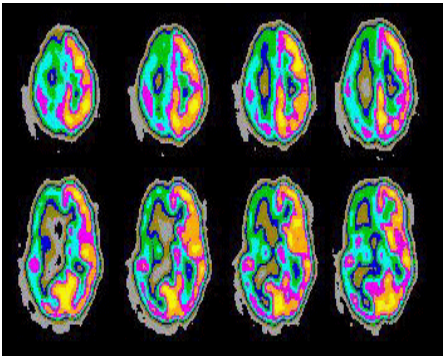

Fig.1 Epileptic seizure caused by pathological conditions

### 3. Nuclear Medicine Imaging

Nuclear medicine imaging shows not only the anatomical structure of an organ or body part, but the function of the organ as well. This functional information can show the organ cell activity very clearly. A small functional change also monitored with the help of this imaging techniques. PET is one of the most popular scanning techniques in current neuroscience research. It is a noninvasive and safe diagnostic procedure. These procedures are limited for a number of diseases, particularly in disease that affect the brain. It is a unique metabolic imaging based on molecular biology. PET scan reveals the cellular level metabolic changes occurring in an organ or tissue before anatomical or structural changes occur. It is an important and unique one often begins with functional changes at the cellular level.

A PET scan can often detect these very early changes whereas a CT or MRI detect changes a little later as the disease begins to cause changes in the structure of organs or tissues. A PET image will display in different levels of positrons according to brightness and color. The images are reconstructed or interpreted by computer analysis and it is used for diagnosing epileptic seizures and assessing its spread.

### 4. PET with epilepsy seizures

In PET scan, a less quantity of radioactive glucose is infected in the body. Brain cells use glucose as fuel, based on this PET examines the absorption of the radioactivity. The level of glucose absorption is used to determine the activity of brain cells. Seizures are known as unsynchronized firing of a cluster of brain cells. Surgical removal of tissues allowed to stop the seizures or minimized significantly. PET is more powerful diagnosis method to identify the source of seizures activity in the brain because it is used non-invasively procedures.

### 5. Technical Approach

Segmentation is a beginning stage for visualization or quantification of medical images for computer aided diagnosis and treatment planning. Image segmentation plays a vital role in automatically delineating the anatomical structures. Various segmentation techniques have proposed in [9]. Most of these techniques are applied to MRI [10] and not to PET images since PET became broadly available recently.

Clustering is used to classify items into similar groups in the process of data mining. It also exploits segmentation which is used for quick bird view for any kind of problem. Symmetry is a pre-attentive feature which improves recognition and reconstruction of shapes and objects. Point symmetry and line symmetry are two keynotes of the geometrical symmetry [11]. In nature symmetry is a basic feature of shapes and objects. Instead of using similarity measure symmetry is also used to find the clusters in the feature space. A non-metric cluster validity measure Sym-index is used in this paper.

$$Sym(K) = \left( \frac{1}{K} \times \frac{1}{\varepsilon_K} \times D_K \right)$$

where

    K is the number of cluster
    $\varepsilon_k$ is the sum of symmetry distance of each cluster
    $D_k$ is the maximum Euclidean distance between two cluster centers among all pairs of centers.

The above validity measure is used for partitioning and to identify convex and non –convex clusters [12].

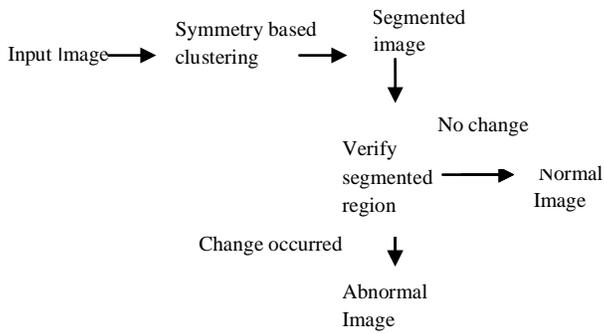

Fig.2 system design for image segmentation based on symmetry

There are many challenges associated with automated detection of brain abnormalities. Fig.2. shows the system design for image segmentation based on symmetrical property.

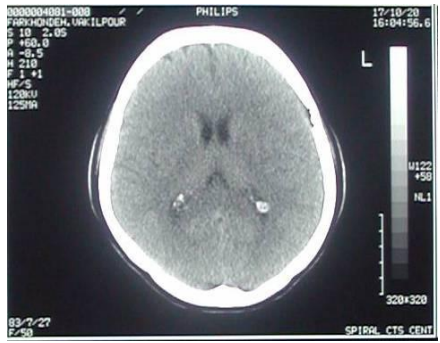

Fig.3 Normal brain picture using PET scan

Normal threshold value of Red, Green and Blue is derived from MIPAV is defined as 85.0 – 170.0. From this threshold value, one can easy to understand the deviation of tissue intensities. Fig. 3 shows the normal brain image using PET scan.

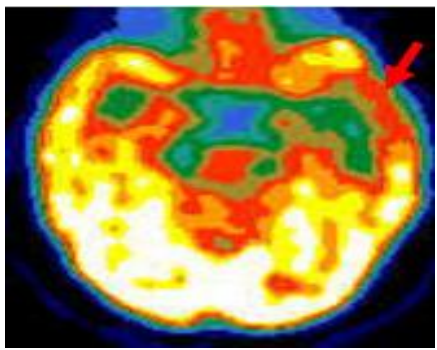

Fig.4 epileptic seizure image using PET scan

Analysis of resulting tissue intensities between each image is noted in Table. 1. It is used to evaluate the performance comparison chart about the abnormalities in specific cerebral regions.

Table 1 Comparison of Pixel Intensities

| Brain image from PET | Red Pixels | Green Pixels | Blue Pixels |
|---|---|---|---|
| Normal image | 134394 | 141086 | 136832 |
| Epileptic seizures | 195426 | 192427 | 193832 |

From the above table pixel intensities variation can be seen in each level. Fig.5 shows the comparisons level a normal as well as abnormal image which is derived from MIPAV tool.

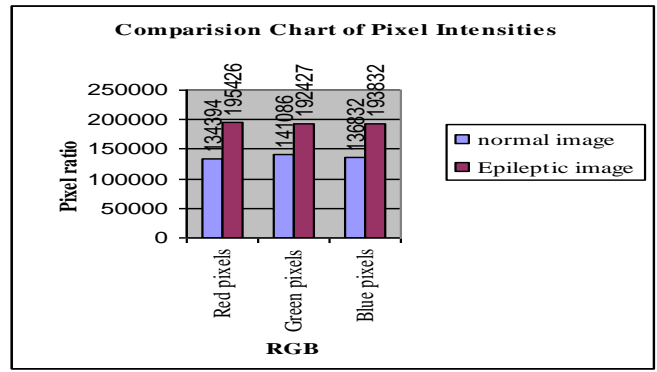

Fig.5 Comparison pixel intensity graph

### 6. Conclusion

This paper provides an automated localization of epileptic seizures in brain from nuclear medicine imaging technique such as PET. Here, MIPAV tool is used to evaluate the comparison result between different kinds of PET images. The datasets used in this paper is derived from University of California, Irvine (UCI), Machine learning repository sample datasets. In future the real sets will be implemented in this approach and to calculate the performance level.